\documentclass[a4paper,conference]{IEEEtran}
\ifCLASSOPTIONcompsoc
  \usepackage[nocompress]{cite}
\else
  \usepackage{cite}
\fi
\ifCLASSINFOpdf
  \usepackage[pdftex]{graphicx}
  \graphicspath{{../pdf/}{../jpeg/}}
  \DeclareGraphicsExtensions{.pdf,.jpeg,.png}
\else
\fi
\usepackage{booktabs}
\usepackage{multirow}
\usepackage{hyperref}
\usepackage{fancyhdr}

\usepackage{textcomp}
\usepackage{fancyhdr}
\usepackage[utf8]{inputenc}
\usepackage{tikz}
\usepackage{textcomp}
\newcommand\copyrighttext{%
  \footnotesize \textcopyright 2020 IEEE. Personal use of this material is permitted.
  Permission from IEEE must be obtained for all other uses, in any current or future
  media, including reprinting/republishing this material for advertising or promotional
  purposes, creating new collective works, for resale or redistribution to servers or
  lists, or reuse of any copyrighted component of this work in other works.}
\newcommand\copyrightnotice{%
\begin{tikzpicture}[remember picture,overlay]
\node[anchor=south,yshift=10pt] at (current page.south) {\fbox{\parbox{\dimexpr\textwidth-\fboxsep-\fboxrule\relax}{\copyrighttext}}};
\end{tikzpicture}%
}
\makeatletter
\@ifundefined{showcaptionsetup}{}{
 \PassOptionsToPackage{caption=false}{subfig}}
\usepackage{subfig}
\makeatother
\usepackage{eso-pic}
\newcommand\AtPageUpperMyright[1]{\AtPageUpperLeft{
 \put(\LenToUnit{0.5\paperwidth},\LenToUnit{-1cm}){
     \parbox{0.5\textwidth}{\raggedleft\fontsize{9}{11}\selectfont #1}}
 }}
\newcommand{\conf}[1]{
\AddToShipoutPictureBG*{
\AtPageUpperMyright{#1}
}
}
\conf{Preprint} 

\begin{document}

%
\title{Handling Concept Drift for Predictions in Business Process Mining}


\author{

\IEEEauthorblockN{Lucas Baier}
\IEEEauthorblockA{
Karlsruhe Institute of Technology (KIT)\\
Karlsruhe, Germany\\
Email: lucas.baier@kit.edu}
\and

\IEEEauthorblockN{Josua Reimold}
\IEEEauthorblockA{
Karlsruhe Institute of Technology (KIT)\\
Karlsruhe, Germany\\
Email: josua.reimold@alumni.kit.edu}

\and
\IEEEauthorblockN{Niklas Kühl}
\IEEEauthorblockA{
Karlsruhe Institute of Technology (KIT)\\
Karlsruhe, Germany\\
Email: niklas.kuehl@kit.edu}
}


%


\maketitle
\copyrightnotice

\begin{abstract}
Predictive services nowadays play an important role across all business sectors. However, deployed machine learning models are challenged by changing data streams over time which is described as concept drift. Prediction quality of models can be largely influenced by this phenomenon. Therefore, concept drift is usually handled by retraining of the model. However, current research lacks a recommendation which data should be selected for the retraining of the machine learning model. Therefore, we systematically analyze different data selection strategies in this work. Subsequently, we instantiate our findings on a use case in process mining which is strongly affected by concept drift. We can show that we can improve accuracy from 0.5400 to 0.7010 with concept drift handling. Furthermore, we depict the effects of the different data selection strategies.
\end{abstract}


%
\IEEEpeerreviewmaketitle

%
%
%
%
%
%
\section{Introduction}
Machine learning plays a major role in the recent developments of artificial intelligence \cite{kuhl2019machine}. It is widely considered to be one of the most disruptive technologies in the last decades. Its fast progress is fueled by both the development of new learning algorithms and the huge availability of low-cost computation and data \cite{Jordan2015}. Machine learning is applied across all sectors and in all functional business areas, such as research and development, marketing or finance \cite{Chen2012a}. Many companies rely on machine learning models for offering new services or for improving their existing ones \cite{Schuritz2016}. As Davenport \cite{Davenport2006} has shown, companies leveraging their data sources achieve a substantial competitive advantage. Especially in the area of services, there seems to be large untapped potential in both, research and practice \cite{Ostrom2015,ching2018ai}.

To address this promising gap, predictive services offer the possibilities to implement machine learning into different application fields \cite{Baier2019}. Typically, techniques of supervised machine learning provide the basis for such predictive services \cite{Jordan2015} which are trained by using historical data of input features and a label. Subsequently, the model is used to continuously compute predictions on a stream of incoming data. However, data streams typically change over time. This is one of the major challenges for applying machine learning in practice \cite{baier2019challenges} since the prediction quality is very sensitive to the input data \cite{Tsymbal2004}. Therefore, the problem of changing data stream over time has been examined under the term “concept drift” \cite{Widmer1996}.

Usual strategies for handling concept drift rely on dedicated drift detection algorithms \cite{Gama2014}. As soon as a drift is detected, the corresponding machine learning model will be retrained. However, it remains an open research question which data instances for the retraining of the machine learning model should be applied (e.g. data before or after the detection). Therefore, we aim at systematically examining the difference between different retraining options which is expressed in RQ1.

\textbf{RQ1.} \textit{Which data should be used for the retraining of a machine learning model when a concept drift is detected?}

Subsequently, we apply our findings of RQ1 in a real-life use case in business process mining, a typical example of a predictive service. Business process management in general, and business process mining in particular, have received a lot of attention recently in top management because it improves decision making in organizations \cite{vanderAalst2016, Rosemann2015}. New applications are extended by the use of predictive analytics \cite{zurMuehlen2015}. Since business processes are inherently dynamic, those new features are largely exposed to concept drift \cite{VanDerAalst2012}. This requires the adaptation of existing methods to ensure their validity over time. Therefore, we want to examine the effects of the different data options on this use case which is regularly confronted with concept drift in the second research question.

\textbf{RQ2.} \textit{What are the effects of the different retraining options in a real-life use case in business process mining?}

The remainder of the paper is structured as follows: \autoref{sec2} presents related work on which we base our research. \autoref{sec3} introduces different aspects which can be considered for the retraining of a machine learning model after detection of a drift. \autoref{sec4} presents the chosen use case as well as the evaluation of the different options discussed in the previous section. The final section discusses our results, describes theoretical and managerial implications, acknowledges limitations and outlines future research.

\section{Related Work}
\label{sec2}
This section gives a brief overview of related work about concept drift as well as its detection. Furthermore, related work regarding process mining is introduced. 

\subsection{Concept Drift}
\label{subsec2.1}
Machine learning can create ongoing value when the corresponding prediction models are deployed in connected information systems and deliver ongoing recommendations on continuous data streams. However, data streams usually change and evolve over time. This is also reflected in changes in the underlying probability distribution or their data structures \cite{Aggarwal2003}. The challenge of changing data streams for machine learning tasks has been described with the term “concept drift” \cite{Widmer1996} in computer science. A concept $p(X, y)$ is defined as the joint probability distribution of a set of input features $X$ and the corresponding label $y$ \cite{Gama2014}. In real applications, concepts often change with time \cite{Tsymbal2004}. This change can be expressed in a mathematical definition as follows \cite{Gama2014}:

\[\exists X: p_{t0} (X,y) \neq p_{t1} (X,y)\]

This definition explains concept drift as the change in the joint probability distribution between two time points $t_0$ and $t_1$. Therefore, machine learning models built on previous data (in $t_0$) might not be suitable for making predictions on new incoming data (in $t_1$). This change requires the frequent adaptation of the prediction approach.

Changes in the incoming data stream can depend on a multitude of different internal or external influences. Usually, it is impossible to measure all of those possible confounding factors in an environment—which is why this information cannot be included in the predictive features of a ML model. Those factors are often considered as “hidden context” of a predictive model \cite{Widmer1996}. Concept drift is usually classified into the following categories \cite{Zliobaite2010}: Abrupt or sudden concept drift where data structures change very quickly (e.g. sensor failure), gradual and incremental concept drift (e.g. change in customers’ buying preferences) or seasonal and reoccurring drifts (e.g. A/C sales in summer). There exists also a more fine-grained taxonomy \cite{Webb2016} which also considers the magnitude of the drift for instance.

A wide variety of approaches for the handling of concept drifts has been proposed \cite{Gama2014}. However, most approaches rely on an explicit drift detection which detects changes in the data distribution and triggers corresponding adaptations. Two of the most popular algorithms are Page-Hinkley \cite{page1954continuous} and ADWIN \cite{bifet2007learning}. Page-Hinkley works by continuously monitoring an input variable (e.g. the input data or the prediction accuracy). As soon as the variable differs significantly from its historical average, a change is flagged. ADWIN, in contrast, is a change detector which relies on two detection windows. As soon as the means of those two windows are distinct enough, a change alert is triggered, and the older window is dropped.

\subsection{Process Mining}
Business process mining is a research discipline that originates from business process modeling and analysis on the one side and data mining on the other side \cite{ManojKumar2015}. The goal of process mining is to discover, monitor and improve operational processes by extracting data from event logs \cite{van2011process}. This way, business processes are analyzed in the way as they are really executed \cite{VanderAalst2004}.  
These event logs can be created by extracting the digital traces of business processes that are stored in today’s information systems, e.g. ERP or CRM systems \cite{van2007business}. The minimum information needed for an event log is therefore a unique CaseID to identify and differentiate each case and an event with relating timestamp to define the activity of the process. This combination is important, so that the real sequence of the events can be ensured. 

Process mining can be differentiated into three types \cite{VanDerAalst2012} where the first type is \textit{discovery}. After extraction of the event logs, a process model can be built. This also allows to understand different variants of business processes \cite{dumas2013fundamentals}. The second type is \textit{conformance}. In this case, existing process models can be compared with an event log of the same process and discrepancies between both can be discovered. The third type relates to \textit{enhancement} where existing process models are extended. This can also refer to operational support where predictions and recommendations based on prediction models from historic information can be used to optimize running cases \cite{VanderAalst2011}. An application could be the prediction of the remaining time of a case \cite{verenich2019survey} or the prediction of the next executed activity in a case \cite{marquez2017predictive}.Furthermore, there are approaches predicting whether a case will be completed \cite{di2017clustering}. With such predictions, the organizational procedures can be optimized, and personnel planning is more accurate. For instance, it can be very valuable for a customer to know the remaining process time of his insurance claim or when his product order will arrive. 

A very important challenge in process mining is the occurrence of concept drift \cite{VanDerAalst2012} which refers to processes that are changing while being analyzed. For instance, the sequence of events can change, e.g. two events that occurred in parallel are now occurring one after another. Processes may change due to a variety of reasons, from seasonal effects over market changes to organizational adjustments. Business processes are inherently dynamic over time and therefore prone to change. Nevertheless, concept drift research in business process mining is rather scarce. Sudden concept drift in process mining, such as rearranging or replacing activities, has been examined \cite{ManojKumar2015}. The authors propose to detect those drifts by computing correlation between event classes. Another approach proposes a framework which computes dedicated features on the event logs and subsequently compares those features over different windows to detect concept drift \cite{bose2013dealing}. In this context, this method to detect drifts is similar to traditional concept drift approaches described in \autoref{subsec2.1}. More advanced options use an adaptive approach based on a Chi-square test which also allows to detect different types of process drift \cite{maaradji2017detecting}. Other research aims at better understanding the type or the degree of change \cite{yeshchenko2019comprehensive} or providing more robustness to process drift detection methods \cite{ostovar2020robust}.

The approaches described above focus on concept drift in the type of event or their order in a process. This is related to the first type of process mining (\textit{discovery}) that focuses on deriving a process model. The ultimate objective of this analysis is to identify and better understand the activities that trigger process drift in the first place. 

\begin{figure*}[htbp]
\centering
\includegraphics[width=4in]{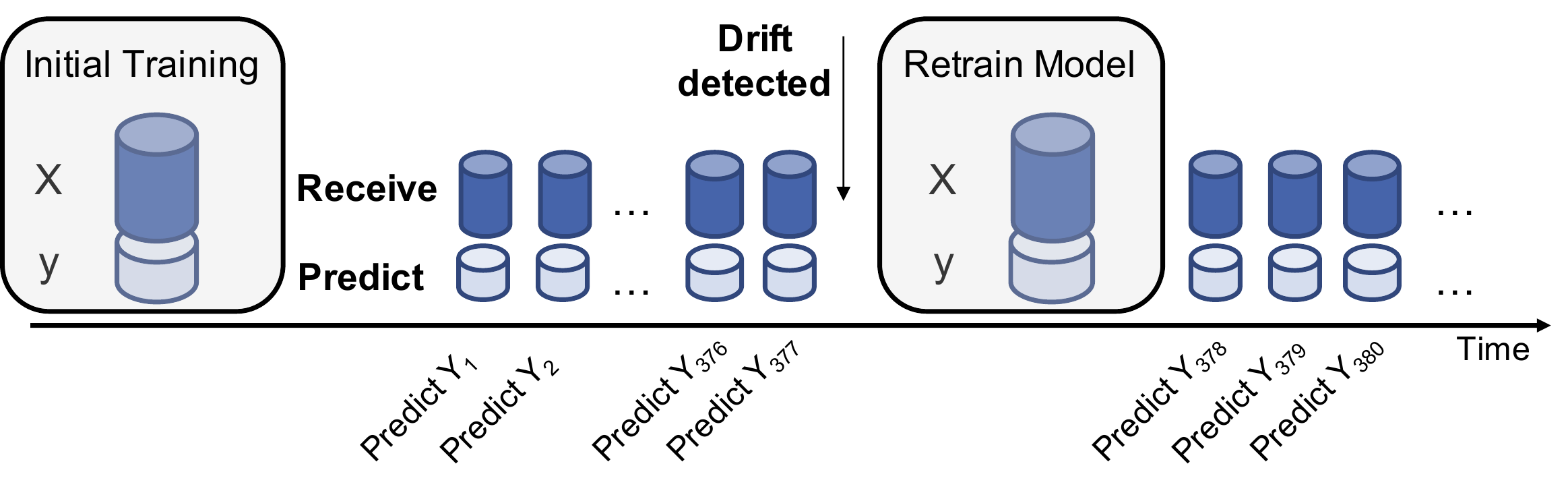}
\caption{Depiction of learning mode retraining}
\label{fig1}
\end{figure*}

However, this analysis does not contain any predictive component. Existing work has not yet considered concept drift in the \textit{enhancement} type of process mining where predictions based on machine learning are computed to optimize operations \cite{VanderAalst2011}. Compared to previous work, this also requires strategies for an adaption of prediction models over time.

\section{Data Selection for Retraining}
\label{sec3}
This section introduces the two different learning modes for machine learning models and provides an overview on which data can be used for the retraining of a model if the training process has to be started from the beginning.

\subsection{Learning mode}
In the context of data streams and ongoing predictions, two learning modes for machine learning models can be differentiated: retraining and incremental learning \cite{Baier2019}. 

The method of retraining is illustrated in \autoref{fig1}. The figure shows that in the beginning the model is trained on an initial batch of data. After the initial model has been trained, new incoming data instances $X$ result in predictions $y$ (e.g., $y_1$ in \autoref{fig1}). This happens iteratively for every new data instance in the data stream until the drift detection method issues an alert which requires an adaptation of the prediction model. Correspondingly, the old model is discarded, and a completely new prediction model is trained which is subsequently applied to every incoming data instance (e.g., the new prediction model after retraining is applied for the first time by predicting $y_{378}$ and the following data instances in \autoref{fig1} ).

Incremental learning, in contrast, works by continuously updating the prediction model. Comparably, the starting model is trained on an initial data set. When new data instances arrive, the model issues a prediction. However, as soon as the true target label of this data instance is known, this information is used to incrementally improve the prediction model. The main advantage of this approach is that every new labeled instance arriving will be used for model improvement and thus, the model automatically adapts to changing concepts. This approach is comparable to a sliding window approach. In general, the incremental updates will not be computed after a single new data instance has arrived but rather after the reception of a small batch of data instances (e.g. 10 or 20). This reduces the computational complexity. Unfortunately, only few machine learning algorithms such as Naïve Bayes, Neural Networks or Hoeffding Trees \cite{Pfahringer2007, vzliobaite2016overview} implement the opportunity for incremental updates.

Despite the continuous updates of the prediction model, this approach might be confronted with degrading performance over time. For instance, the incremental updates of the model cannot adapt to very quick changes which occur during sudden concept drifts. In this case, it might be also necessary to discard the current model and train a new model. This would depict a combination of both learning modes retraining and incremental updates.

\subsection{Data Selection for Retraining of the Machine Learning Model}

\begin{figure*}[htbp]
\centering
\includegraphics[width=4in]{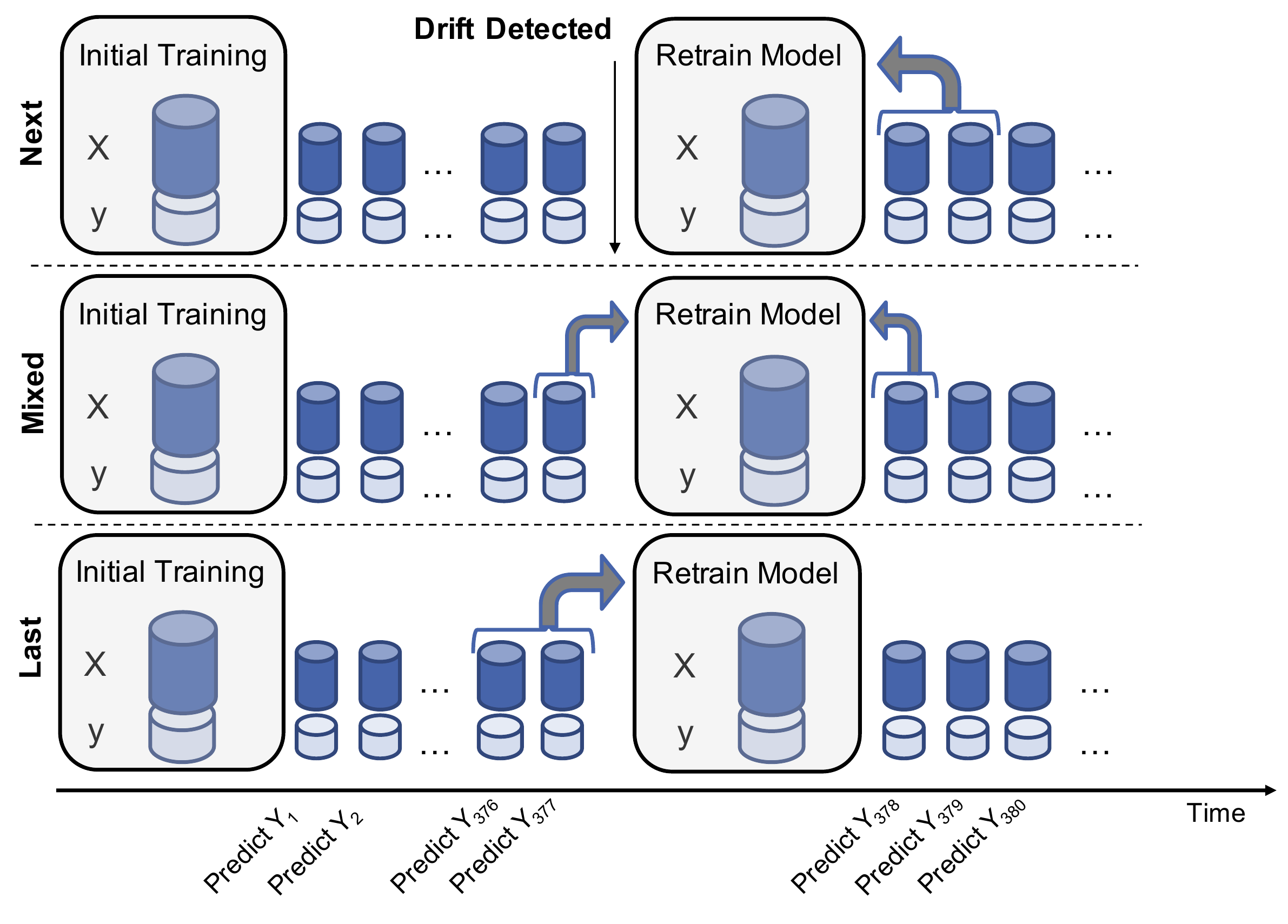}
\caption{Three different approaches for retraining of model}
\label{fig2}
\end{figure*}

In case of concept drift, the previous model will be discarded, and a new model is trained as depicted in \autoref{fig1}. However, when implementing this approach, we need to select the data that is used for the retraining of the machine learning model. So far, literature does not provide any knowledge on which data of the data stream should be used for the retraining of the prediction model. Therefore, we implement and evaluate three different data selection strategies which we call \textit{next}, \textit{mixed} and \textit{last}. The difference between these approaches is depicted in \autoref{fig2}.

The approach \textit{next} is displayed in the upper part of \autoref{fig2}. As soon as a concept drift is detected, the model collects the next batch of instances with corresponding labels (e.g. two new data instances in the figure). When this next batch is complete, the retraining is started and subsequently the new model is applied. This also means that the previous model is used to predict the next batch after the concept drift since it is also necessary to issue predictions for those instances (and the new model has not been learned yet). The intuition guiding this approach is that data following a concept drift, complies with the new concept and is therefore an optimal basis for a new model. 

The other approaches \textit{mixed} and \textit{last} are also displayed in \autoref{fig2}. In case of the \textit{mixed} approach, the model retraining relies on data from before and also after the detection. Compared to the first approach (\textit{next}), the new model can be applied faster since it requires less data after the concept drift. The \textit{last} approach entirely relies on data which was acquired before the concept drift detection alert. This means that the new prediction model will be applied right on the next data instances after the detection of a drift. This approach might work well because drift detection algorithms usually work with a slight delay. Therefore, the data batch before the alert might already belong to the new concept.

During the application of our use case, we aim to systematically test all three approaches in order to quantify the differences between those and also to give recommendations for future implementations.

\section{Use Case in Process Mining}
\label{sec4}

\begin{figure}[htbp]
\centering
\includegraphics[width=3.2in]{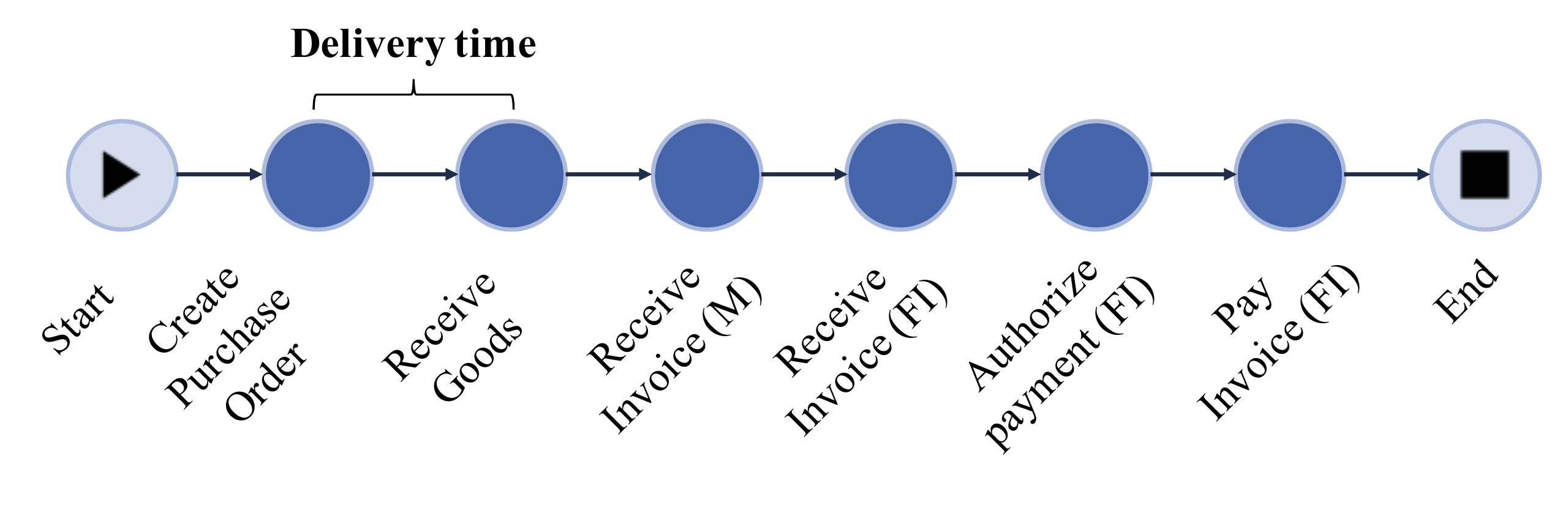}
\caption{Typical process variant for a P2P process}
\label{fig3}
\end{figure}

A process mining solution provider gives us access to a data set of the purchase to pay (P2P) service process of a large German company. This process contains all activities related to the procurement of a product or service. A simplified P2P process starts with the creation of a purchase order and is followed by the reception of the respective goods by the logistics department and the invoice which is then processed over various financial departments in the company. An exemplary process of this P2P process can be seen in \autoref{fig3}. 

In this use case, we want to predict the throughput time or delivery time (marked in bold) between the creation time of the purchase order and the reception of the goods. This information is quite important for the company since all subsequent process steps such as production can be optimized, and significant cost savings can be realized. The data is extracted from the business intelligence platform Qlik and then preprocessed in Python. The foundation of the data set is an event log that is enriched with numerous additional attributes to fully describe the process. The attributes are anonymized and transformed to ensure that the data is not retraceable. In total, we receive data about 70,774 purchase transactions from 2016 until 2018 which we can use to train and evaluate the machine learning approach. Importantly, those transactions are displayed in chronological order, which is a necessary prerequisite for an analysis of concept drift over time.

We use the package scikit-mulitflow \cite{Montiel2018} as the basis of our analysis since it extends the machine learning package scikit-learn with a stream data framework. It allows to process data sets and simulate them as a data stream. Furthermore, different concept drift detectors are implemented and can be evaluated. We extend the package by implementing the different training modes (\textit{last}, \textit{mixed}, \textit{next}) which we discussed in \autoref{sec3}.

\subsection{Data Analysis}

We first perform an exploratory data analysis to analyze the available features and build a predictive model that can be used for the analysis of concept drift in process mining. \autoref{table1} gives an overview on available features of the data set. Categorical features are one-hot-encoded for the subsequent data processing. \textit{Material class} refers to the product category of the purchased product. Regarding this feature, we only use the first four numbers of the material class in order to reduce the number of different categories resulting in 123 different categories in total. Furthermore, we have information about the \textit{purchase order value}. The purchase order value is an important feature for our endeavor since it is a clear indicator of the relevance of the respective purchase order for the company. However, the distribution of the order value is highly skewed which might pose a problem for the prediction model. Therefore, the values are transformed with a Box-Cox transformation \cite{Box1964} into a gaussian distribution.

\begin{table}[htbp]
\caption{Overview of predictive features}
\label{table1}
\centering
\begin{tabular}{l l l}
\toprule
Feature & Type & Number of items /\\
& & Range of values\\
\midrule
Material class & categorical & 123\\
Document type & categorical & 7\\
Plant code & categorical & 4\\
Purchase order value & numerical & 1 – 458,079\\
Supplier & categorical & 799\\
Bank country & categorical & 18\\
Supplier country & categorical & 14\\
Purchasing group & categorical & 75\\
\midrule
Throughput time [h] (Target) & numerical & 1 – 120,000\\
\bottomrule
\end{tabular}
\end{table}

Other features included in the data set are the \textit{country of the bank} were the payment is executed and the \textit{document type} of the purchase order. The \textit{document type} includes information about different ways to create a purchase order: e.g., the order is created manually by an employee in the purchasing department or is based on existing long-time contracts. Other options include the automatic creation by an MRP-system. The \textit{country of the supplier} is also relevant for the analysis. Obviously, a purchasing process requires more time if the supplier is located in another country because this leads to additional steps during the sales process such as customs papers, currency conversion or additional insurance of the transport. The feature \textit{plant code} stores information about the plant which initiated the purchase process. \textit{Purchasing group} is the department or group at which the purchase order is created and processed. Furthermore, we also have information about the \textit{supplier} itself who is distributing the requested product.

The \textit{target variable} in this use case is the \textit{throughput time} or delivery time of a purchase order. This refers to the amount of time between the first two steps depicted in \autoref{fig3}. By considering the delivery time, we ensure that the start of the purchase process is considered as well as the most important event for production and workforce scheduling, namely the arrival of the ordered goods. The prediction of the estimated arrival time of a product is important because planning processes can be optimized with this information. This might result in significant cost savings as well as the minimization of production time due to the optimization of waiting time.

\begin{figure}[htbp]
\centering
\includegraphics[width=3.2in]{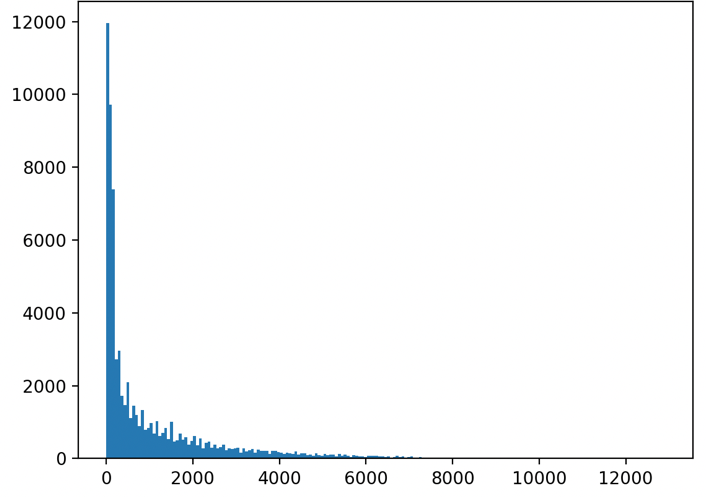}
\caption{Histogram of the throughput time [h]}
\label{fig4}
\end{figure}

A histogram of the throughput time can be seen in \autoref{fig4}. For approximately 50\% of all purchase orders, respective products and goods are received within 14 days ($<$336h). Regarding the remaining purchase orders, another 25\% of those have a delivery time within 60 days. The other purchase orders even have a larger delivery time, up to 537 days. 

Due to the challenging distribution of the target variable, we transform the use case into a multi-class classification problem. Although this leads to an abstraction and loss of information, this step is meaningful for an initial analysis of the use case. To transform the target variable, all purchase orders are divided into three equally sized classes of throughput times as can be seen in \autoref{table2}. Therefore, the first class contains purchase orders with a delivery time of up to 6 days. The second class contains purchase orders with a delivery time between 7 and 39 days and the last class contains all cases for which the delivery takes more than 40 days. We train a machine learning model which predicts whether a purchase order will belong to the short, medium or large throughput time class. 
\begin{table}[htbp]
\caption{Overview of multi-class target variable}
\label{table2}
\centering
\begin{tabular}{l l l l}
\toprule
& Short time & Medium time & Large time\\
\midrule
Delivery time & 0 – 6 days & 7 – 39 days & $>$ 39 days\\
\bottomrule
\end{tabular}
\end{table}

\subsection{Evaluation of prediction}
We first perform a pretest with various machine learning algorithms in their standard parameter configuration \cite{Pedregosa2011}: Naïve Bayes, Neural Network, Support Vector Machine and Decision Tree. The results depicted in \autoref{table3} are computed by performing a 70\%-30\% train-test-split on the first 2,000 data instances. We assume that those data instances all belong to the same concept as there is no significant change observable in the input data. Therefore, we can safely apply the machine learning algorithms without considering and handling concept drift. Note that prediction performance on later parts of the data set might be lower due to the challenges induced by concept drift.

\begin{table}[htbp]
\caption{Pretest with different models on subset of data}
\label{table3}
\centering
\begin{tabular}{l|c}
\toprule
Model & Accuracy\\
\midrule
Naïve Bayes & 0.767\\
Neural Network & 0.805\\
Support Vector Machine & 0.697\\
ecision Tree & 0.740\\
\bottomrule
\end{tabular}
\end{table}

Naïve Bayes, Neural Networks and Decision Trees all achieve similar accuracy values. We choose Naïve Bayes classifier as the prediction algorithm which is due to two reasons: First, Naïve Bayes implements incremental learning which allows incremental uptates of the prediction model. Second, computational complexity of Naïve Bayes is rather low compared to other machine learning algorithms which allows frequent retraining of the model without the necessity for a large computational infrastructure. 

 Our work mainly focuses on the quantification and handling of concept drift. However, we do not have any knowledge whether there are any drifts at all in the data set or at which point in time they are occurring. Therefore, first of all, we analyze the impact of concept drift by applying a Naïve Bayes classification without any concept drift detection method---called ``static model''---to the entire data set of 70,774 data instances. Subsequently, we apply Naïve Bayes classifier in combination with a Page-Hinkley test and ADWIN as drift detection methods. As evaluation metric, we use the accuracy by measuring how often the algorithm predicts the appropriate throughput time class. This metric is chosen since the instances are distributed equally over all three target classes. 
 
\begin{figure}[htbp]
\centering
\includegraphics[width=3.5in]{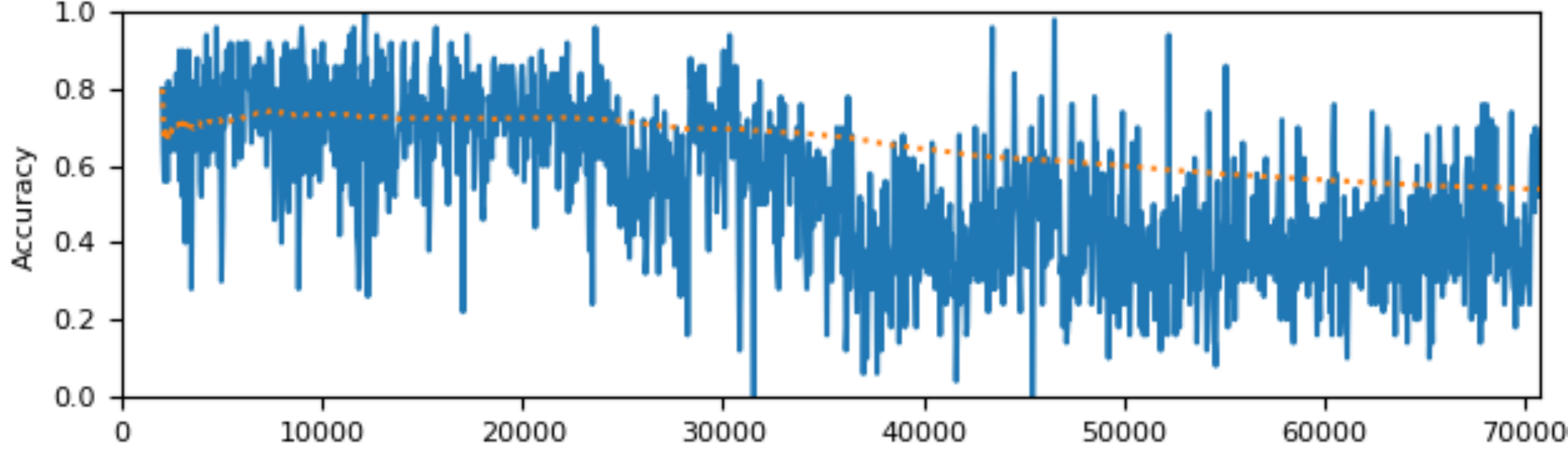}
\caption{Accuracy of Naïve Bayes without retraining and no drift detection method}
\label{fig5}
\end{figure}

The course of the accuracy of the static model without concept drift detection and incremental learning can be seen in \autoref{fig5}. The first 2,000 data instances are used for the initial training. Subsequently, we compute the first predictions and the accuracy level moves at around 0.7. Then, there is a first drop in accuracy after approximately 25,000 instances. However, the prediction performance recovers to around 0.7 shortly after. Subsequently, after approximately 35,000 instances, the prediction quality of the model decreases significantly. Supposedly, a concept drift has occurred because the model that is only trained on an initial data batch does not issue any useful prediction anymore. The accuracy over all predictions reaches 0.5400.

\begin{figure}[htbp]
\centering
\includegraphics[width=3.5in]{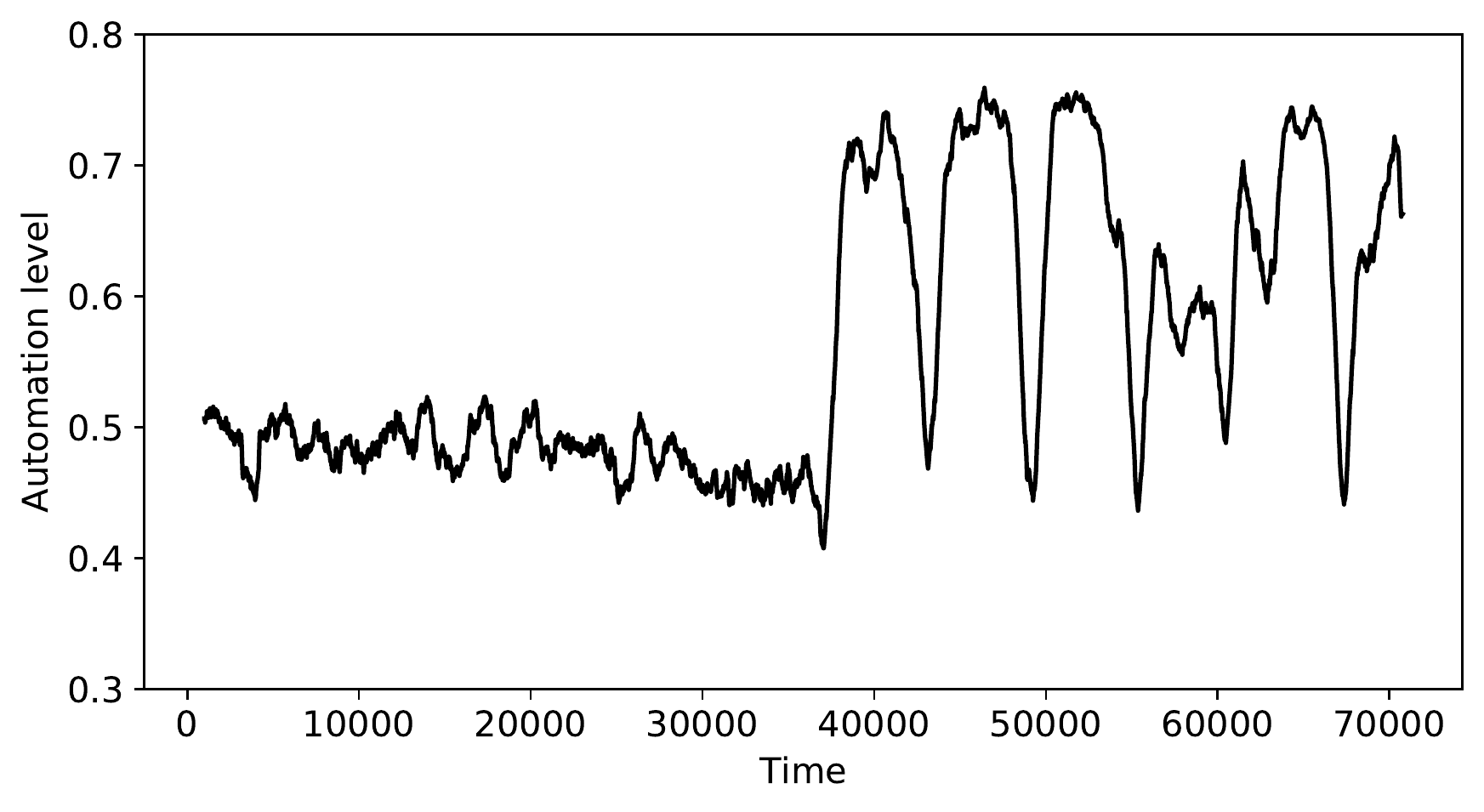}
\caption{Rolling mean (window 1000) of feature \textit{automation}}
\label{fig6}
\end{figure}

As usual, it is difficult to determine the underlying reasons for this concept drift with certainty \cite{vzliobaite2016overview}. However, after a thorough analysis of additional data—which is not available the moment when the prediction is computed—we identify a possible explanation. The feature \textit{automation} contains information about the percentage of process steps in the entire P2P process which are executed automatically by corresponding information systems, while the other steps are executed manually. Thereby, the feature \textit{automation} contains information about the level of automation in all processes. In order to analyze the development of this feature over time, we compute and plot a rolling mean ($window=1000$) of this feature which is depicted in \autoref{fig6}.

At first, the rate of automation is rather stable before it rises abruptly and then fluctuates at a higher level. This plot clearly indicates on how the automation rate in the organization increases over time and thus, this may be one of the causes for concept drift and according changes in product delivery times. The sudden rise in automation maps rather well to the decrease in prediction accuracy in \autoref{fig5}. Relating to \autoref{sec2}, this abrupt change can be seen as a sudden concept drift. Since this feature is not known at the time of prediction, it can be interpreted as a hidden context influencing the prediction.

\begin{figure}[htbp]
\centering
\includegraphics[width=3.5in]{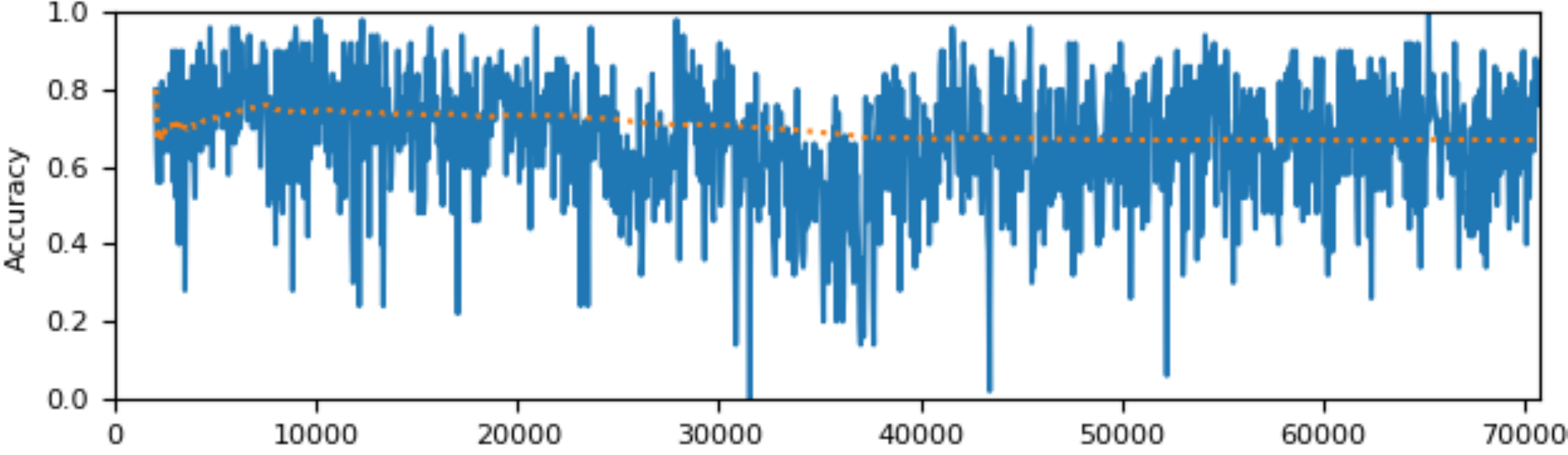}
\caption{Accuracy of NaïveBayes with Page-Hinkley}
\label{fig7}
\end{figure}

Due to the detected drift, we apply a Page-Hinkley test as concept drift detection method in combination with the Naïve Bayes classifier. In case of drift, the model is retrained. The course of the accuracy of the model can be seen in \autoref{fig7}.

At the beginning, the figure looks similar to the model without drift detection (\autoref{fig5} above). After approximately 35,000 instances, this model performs better because the drift is detected, and a retraining of the Naïve Bayes is triggered. The accuracy rises again and then stays same level with its corresponding fluctuations leading to an overall accuracy of 0.6732 (see \autoref{table4}). This is equivalent to a performance increase of 24\%. Furthermore, we extend this approach by activating incremental learning. This means that the model is constantly updated with new training data after it has issued prediction for those data. The application of incremental learning alone leads to a performance of 0.6717. With both retraining and incremental learning, the overall prediction accuracy reaches 0.6938.

\begin{table}[htbp]
\caption{Performance of different data selection strategies on entire process mining data set}
\label{table4}
\centering
\begin{tabular}{c|c|c|c}
\toprule
\multirow{2}*{Change detection} & Incremental & \multirow{2}*{Accuracy} & Performance\\
& learning & & increase\\
\midrule
None (baseline) & No & 0.5400 & –\\
None & Yes & 0.6717 & 24.39\%\\
Yes (Page-Hinkley) & No & 0.6732 & 24.67\%\\
Yes (Page-Hinkley) & Yes & 0.6938 & 28.48\%\\
\bottomrule
\end{tabular}
\end{table}

We perform a grid search on the first 10,000 data instances in order to optimize the parameters of the drift detection method ADWIN ($\delta = 0.001$) and Page-Hinkley ($\lambda = 0.6$). With those parameters, we evaluate the different data selection strategies as discussed in \autoref{sec3}. \autoref{table5} depicts the accuracy score of a Naïve Bayes classifier with incremental learning in combination with a Page-Hinkley test or ADWIN as drift detection. Furthermore, we examine the influence of four different batch sizes (500, 1000, 2000, 5000) on the overall prediction accuracy. This refers to the amount of data instances which are provided to the model in case of retraining. The best results are marked in bold in \autoref{table5}.

As depicted in the table, the data selection strategy \textit{last} performs always best. For our use case, Page-Hinkley appears to be the more suitable drift detector resulting in higher performance. Interestingly, the prediction accuracy decreases with increasing batch size which might indicate that the approach does not adapt fast enough with larger batches for retraining. Furthermore, the performance difference between the different data selection strategies also rises with the size of the batches. For instance, the difference between \textit{last} and \textit{next} for Page-Hinkley with batch size 500 equals 0.0073 in comparison to 0.0164 for Page-Hinkley with batch size 5000.

\begin{table}[htbp]
\caption{Performance of different data selection strategies on process mining data set}
\label{table5}
\centering
\begin{tabular}{l|c|c c c}
\toprule
Change detection & Incremental & \multirow{2}*{Last} & \multirow{2}*{Mixed} & \multirow{2}*{Next}\\
(batch size) & learning & &\\
\midrule
Page-Hinkley (500) & Yes & \textbf{0.7010} & 0.6961 & 0.6937\\
Page-Hinkley (1000) & Yes & \textbf{0.6965} & 0.6920 & 0.6903\\
Page-Hinkley (2000) & Yes & \textbf{0.6938} & 0.6845 & 0.6821\\
Page-Hinkley (5000) & Yes & \textbf{0.6842} & 0.6757 & 0.6678\\
\midrule
ADWIN (500) & Yes & \textbf{0.6856} & 0.6849 & 0.6843\\
ADWIN (1000) & Yes & \textbf{0.6854} & 0.6838 & 0.6825\\
ADWIN (2000) & Yes & \textbf{0.6803} & 0.6775 & 0.6750\\
ADWIN (5000) & Yes & \textbf{0.6758} & 0.6704 & 0.6675\\
\bottomrule
\end{tabular}
\end{table}

In general, the evaluation section clearly shows how the prediction performance can be increased by implementing drift handling strategies. Both, incremental learning as well as drift detection with retraining, have significant influence on the accuracy. Best results are achieved with the combination of both approaches.

\section{Conclusion}
Process mining relies more and more on techniques of machine learning. This work explores the challenge of concept drift for ongoing value creation in process mining. Specifically, we apply a concept drift detection algorithm on a use case which aims at predicting the delivery time for all purchase orders of a company. With this information, the company can optimize its internal service processes. We can show that concept drift handling significantly outperforms a static model in the given use case. Best results are achieved by combining incremental learning with retraining in case of concept drift. Regarding the best training data selection strategy for retraining, the \textit{last} approach appears to be the best performing option. This means that data scientists should rely on the last collected data batch for the retraining of the prediction model.

The contribution of this paper is twofold. First, we systematically explain and depict the options for training data selection for the retraining of machine learning models in case of concept drift. Second, we apply and evaluate those options in a real-life use case in process mining where we can measure a significant increase in prediction performance from 0.5400 to 0.7010. Regarding the managerial implication, this work clearly shows the importance of a continuous monitoring and adaptation scheme of predictive services in operation. Otherwise, they can quickly lose their validity and corresponding service offerings will not deliver expected benefits.

However, more research is required to understand the full effects of concept drift and the best strategies to deal with this problem. This work only describes and evaluates three options for the training data selection in case of retraining. Future work needs to evaluate more sophisticated approaches. Additional limitations regarding the use case arise through the transformation of the target variable from a regression problem into a multi-class classification problem. Furthermore, we only evaluate the data selection on one use case. More general recommendations could be derived by applying those options onto more use cases and benchmark data sets. 

This paper clearly shows the importance of constant monitoring of predictive services for the detection of concept drifts. Frequent retraining and adaptations of a machine learning model are necessary requirements to keep and guarantee a high prediction performance. If practitioners consequently implement necessary monitoring activities, the economic benefits of predictive services and supervised machine learning solutions can still even be increased.

\bibliographystyle{IEEEtran}

\end{document}